\pdfoutput=1

\documentclass[11pt]{article}

\usepackage[]{EMNLP2023}

\usepackage{times}
\usepackage{latexsym}

\usepackage[T1]{fontenc}

\usepackage[utf8]{inputenc}

\usepackage{microtype}

\usepackage{inconsolata}
\usepackage{graphicx}

\usepackage{listings}
\usepackage{color}

\definecolor{dkgreen}{rgb}{0,0.6,0}
\definecolor{gray}{rgb}{0.5,0.5,0.5}
\definecolor{mauve}{rgb}{0.58,0,0.82}

\title{Using Zero-shot Prompting in the Automatic Creation and Expansion of Topic Taxonomies for Tagging Retail Banking Transactions}

\author{Daniel de S. Moraes$^1$,~ Pedro T. C. Santos$^1$,~ Polyana B. da Costa$^1$, \\ 
    {\bf Matheus A. S. Pinto$^1$,} {\bf~ Ivan de J. P. Pinto$^1$,} {\bf~ Álvaro M. G. da Veiga$^1$,} {\bf~ Sergio Colcher$^1$,} \\
    {\bf Antonio J. G. Busson$^2$,} {\bf~ Rafael H. Rocha$^2$,} {\bf~ Rennan Gaio$^2$,} 
    {\bf Rafael Miceli$^2$,} \\ {\bf Gabriela Tourinho $^2$,} {\bf Marcos Rabaioli $^2$,} {\bf~ Leandro Santos$^2$,} {\bf~ Fellipe Marques$^2$,} {\bf~ David Favaro$^2$} \\ \\
        $^1$Pontifical Catholic University of Rio de Janeiro \\
        $^2$BTG Pactual}

\begin{document}
\maketitle
\begin{abstract}
This work presents an unsupervised method for automatically constructing and expanding topic taxonomies using instruction-based fine-tuned LLMs (Large Language Models). We apply topic modeling and keyword extraction techniques to create initial topic taxonomies and LLMs to post-process the resulting terms and create a hierarchy. To expand an existing taxonomy with new terms, we use zero-shot prompting to find out where to add new nodes, which, to our knowledge, is the first work to present such an approach to taxonomy tasks. We use the resulting taxonomies to assign tags that characterize merchants from a retail bank dataset. To evaluate our work, we asked 12 volunteers to answer a two-part form in which we first assessed the quality of the taxonomies created and then the tags assigned to merchants based on that taxonomy. The evaluation revealed a coherence rate exceeding 90\% for the chosen taxonomies. The taxonomies' expansion with LLMs also showed exciting results for parent node prediction, with an f1-score above 70\% in our taxonomies.

\end{abstract}

\section{Introduction}
\label{sec:intro}

Many recent works have been devoted to financial transaction classification/characterization using Machine Learning-based methods. For instance, hierarchically classifying financial transactions was investigated by \citealp{vollset2017making} and \citealp{busson2023hierarchical}. They both use a set of predefined categories and subcategories that cover (and presumably describe) the universe of consuming types and transactions. 

However, using such a small predefined (and also statically defined) set of classes limits the ability to extend classification capabilities based on the user's experiences or when facing new categories not previously designed. 

In this context, to expand the possible set of classes/tags to label a transaction, we developed an unsupervised method based on \emph{topic taxonomies}. Taxonomies are very useful in the structural and semantic analyses of topics and textual data. However, creating and maintaining them is often costly and challenging to scale manually. Therefore, recent works have tackled the automatic creation and expansion of \emph{topic taxonomies}, in which each node in a hierarchy represents a conceptual topic composed of semantically coherent terms.

We present an unsupervised method for automatically constructing and expanding topic taxonomies with instruction-based LLMs (Large Language Models). To build an initial taxonomy, we apply topic modeling, keyword extraction techniques, and LLMs to post-process the resulting terms, create a hierarchy, and add new terms to an existing taxonomy. Since the taxonomies are derived from a corpus of unstructured texts describing niches of consuming habits, we opted to investigate the use of LLMs in our approach. LLMs are often pre-trained on a large corpus of text, allowing them to learn contextual representations that capture the intricacies of human language. 

We applied our method in an enriched retail bank transactions dataset (with scraped data from food and shopping companies) and evaluated the resulting taxonomies quantitatively. The generated tags of our topic taxonomies are then assigned to the bank clients' transactions to characterize the establishments in each transaction. In total, 58 topic taxonomies were created for the \emph{Food} taxonomy and 6 for the \emph{Shopping} taxonomy.

A two-step qualitative evaluation is conducted over a limited part of the taxonomies. For this evaluation, we selected the topic taxonomies with the highest number of terms in each taxonomy ("Brazilian Cuisine" from \emph{Food} and  "Clothing and Accessories" from \emph{Shopping}). We asked volunteers to answer a two-part form. First, we assess the quality of the taxonomies created and then the quality of the tags assigned to label merchants. The evaluation showed an average coherence above 90\%.

The expansion of taxonomies with Commercial LLMs like Gemini~\cite{team2023gemini} and GPT-4~\cite{openai2023gpt4}, alongside open-source LLM options such as LLaMA-Alpaca (7B)~\cite{touvron2023llama}, Phi-2\footnote{\url{https://huggingface.co/microsoft/phi-2}}, and Mixtral 8x7B~\cite{jiang2024mixtral}, is evaluated across four taxonomies from the SemEval dataset and our taxonomies. Gemini achieved the best results, with f1-scores of 89\% and 70\% for \emph{Food} and \emph{Shopping} taxonomies, respectively.

\section{Related Work and Background}

Taxonomies represent the structure behind a collection of documents, organizing the hierarchical relationships between terms in a tree structure \cite{nikishina2020russe}. They play an essential part in the structural and semantic analysis of textual data, providing valuable content for many applications that involve information retrieval and filtering, such as web searching, recommendation systems, classification, and question answering. 

Since creating and maintaining taxonomies is a costly task, often difficult to scale if done manually, methods that automatically construct and update them are desirable. Early works on automatic taxonomy creation focused on building hypernym-hyponym taxonomies, where each pair of terms expresses an `is-a' relationship \cite{snow2004learning}. More recent works have tackled the automatic creation of other taxonomies, such as topic taxonomies. In a taxonomy, each node represents a conceptual topic composed of semantically coherent terms. 

In this context, \citealp{taxogen} developed TaxoGen, an unsupervised method for constructing topic taxonomies. Taxogen uses the SkipGram model from an input text corpus to embed all the concept terms into a latent space that captures their semantics. In this space, the authors applied a clustering method to construct a hierarchy recursively based on a variation of the spherical K-means algorithm. 

Another work that focuses on topic taxonomies is TaxoCom, a framework for automatic taxonomy expansion \cite{taxocom}. TaxoCom is a hierarchical topic discovery framework that recursively expands an initial taxonomy by discovering new sub-topics. TaxoCom uses locally discriminative embeddings and adaptive clustering, resulting in a low-dimensional embedding space that effectively encodes the textual similarity between terms. Regarding the automatic expansion of taxonomies, one important related example is Musubu \cite{musubu}, a framework for low-resource taxonomy enrichment that uses a Language Model (LM) as a knowledge base to infer term relationships. 

Regarding LMs, recent breakthroughs in deep learning, particularly with the advent of Transformer models, have revolutionized the field \cite{vaswani2017attention}. Pre-trained LMs, such as OpenAI's GPT \cite{radford2019language} and Google's BERT \cite{devlin2018bert}, have achieved remarkable performance across a wide range of NLP tasks, including language generation, sentiment analysis, and named entity recognition. Lately, LLMs have garnered significant attention for their exceptional performance in such tasks. LLMs, such as GPT-3 \cite{brown2020language} and LLAMA \cite{touvron2023llama}, are characterized by their massive scale, comprising billions of parameters and being trained on vast data.

To use LLMs for specific purposes, a highly effective approach is to fine-tune them on task-specific data. In this context,  InstructGPT proposes a method called RLHF (Reinforcement Learning from Human Feedback), which involves fine-tuning large models using PPO (Proximal Policy Optimization) based on a reward model trained to align the models with human preferences ~\cite{openai2023gpt4, ouyang2022training}. 

Along with fine-tuning, Prompt Tuning is also applied to enhance the performance and adaptability of LLMs in specific tasks or domains. Prompt Tuning involves designing and optimizing prompts or instruction formats for LLMs \cite{brown2020language}. Techniques such as template-based prompts, rule-based prompts, and cloze-style prompts have been explored to guide LLMs toward desired responses. 


\section{Dataset Construction}


This work used a private dataset of a retail bank's consuming transactions. Each transaction has a merchant name representing the business where that purchase occurred and macro and micro categories that characterize that business \cite{busson2023hierarchical}. We chose two macro-categories from this dataset: \emph{Food} and \emph{Shopping}, using the top 50,000 businesses with the highest number of transactions for each category.

With such little information, assigning tags that identify a transaction in more detail is difficult. To gather more data that could be helpful in the next steps of our method, we performed a data enrichment step in the dataset through web scraping. We searched for establishments in each macro category with descriptions of their activities. For the \emph{Food} macro category, the search can be performed on specific platforms for restaurants and food delivery. As for the \emph{Shopping} macro category, we searched for establishment/business descriptions directly on internet indexing and search tools.

Specifically for the \emph{Shopping} macro-category, we collected descriptions using the merchant names and micro categories as the query in the search tools. As for the ones in the \emph{Food} category, we achieved coverage in collecting descriptions in a specific food delivery platform. Additionally, for the \emph{Food} category, we collected data on the type of food provided at those establishments (e.g., Japanese, Mexican, Brazilian, etc.). 


\section{Taxonomy Construction}

For the automatic creation of topic taxonomies for \emph{Food} and \emph{Shopping} businesses, we developed a 3-step method. At first, we preprocess the descriptions in our dataset to keep only relevant parts of the text. Subsequently, we apply two techniques to select candidate terms for the topic taxonomies: Keyword extraction and Topic Modeling. 

For post-processing, we use an LLM to refine the result of each step, identifying unrelated terms. Lastly, we use an LLM again to organize the final terms into hierarchies to configure the topic taxonomies. 

\subsection{Preprocessing}
\label{preprocessing}

We applied a few NLP techniques to refine the businesses' descriptions in our dataset. At first, we remove stop words to eliminate commonly used words that do not carry significant meaning in our contexts. Then, to retain only the most relevant portions of the descriptions, we employ part-of-speech (POS) tagging to identify and exclude words that belong to specific POS categories. The list of POS tag categories that were removed includes ADV, CCONJ, ADP, AUX, CONJ, DET, INTJ, PART, PRON, PUNCT, SYM, SCONJ, ADJ, VERB, PROPN.\footnote{https://spacy.io/usage/linguistic-features\#pos-tagging}

After this initial preprocessing step, we run the first iteration of the candidate term selection part to build a filter of generic words, not to create topic taxonomies yet. For this step, we use the entire corpus of descriptions for each macro category, resulting in two corpora (\emph{Food} and \emph{Shopping}). For each micro category in the macro categories' corpora, we use Keyword Extraction and Topic Modeling to gather candidate terms for the filter, combining the results of both techniques in a list. Then, We use an LLM to remove the terms it identifies as unrelated to the main topic (each micro category) from the list. The prompt that we used for requesting this separation is illustrated below.

\begin{lstlisting}[caption=Prompt for separating candidate terms related to the type of establishment,captionpos=b]
prompt= "Given the terms in the following list: "+ <words_list> +". Separate them into two groups. In group 1 the terms with no relation to the topic "+ <type> +". And in group 2 the terms that are related."
\end{lstlisting}

By using this prompt, we ensure that the model's response is consistently formatted according to the pattern described in it, facilitating the processing of the resulting string. Once we complete one iteration of this method for each macro category in our dataset, we add the words of group 2 to the corresponding list of generic words. We apply the corresponding filter of generic words for each macro category corpus, resulting in the final preprocessed corpus.

\subsection{Candidate Terms Selection}\label{taxonomyC}

For this part of our method, we use each preprocessed corpus separately. For the \emph{Food} corpus, we group the descriptions based on their micro-categories, creating 58 sub-corpus specific to that domain. We have six micro categories for the \emph{Shopping} corpus, resulting in 6 specific sub-corpus. The candidate terms selection methods are applied to each sub-corpus, creating topic taxonomies where the main topic is the micro category.  

\subsubsection{Keyword Extraction}
The first approach to candidate term selection was to use an unsupervised keyword selection method called 
Yake! \cite{campos2020yake}. This method is based on statistical text features extracted from single documents to select the most relevant keywords from that text. It does not require training on a document set and is not dependent on dictionaries, text size, language, domain, or external corpora.

Yake! allows for the specification of parameters such as the language of the text, the maximum size of the n-grams being sought, and others. In our method, we customized only the language to Portuguese, and the maximum number of keywords sought for each set of descriptions was 30 words.

After extracting the keywords from each group of descriptions, we obtained a total set of $N$ candidate terms. However, these terms are further filtered using an LLM, where we ask it to separate the terms related to the main topic from those unrelated, as explained earlier in subsection \ref{preprocessing}.

\subsubsection{Topic Modeling}

Our second approach to collecting initial topics and candidate terms was Topic Modeling. We applied the Latent Dirichlet Allocation algorithm \cite{lda}, available at the Gensim Library.\footnote{https://pypi.org/project/gensim/}

We construct a dictionary for each macro-category corpus in our macro-categories corpora by extracting unique tokens and bigrams. After a few empirical tests, we set the minimum frequency of a bigram to 20 occurrences. Since some corpora have a minimal number of tokens (the micro category "Greek Cuisine" from the \emph{Food} macro category has only five stores marked as such, with a corpus of only 127 tokens), we had to set a reasonably small number so that smaller corpora could also have a few bigrams. With the resulting dictionary of tokens, the LDA algorithm was applied. Three main parameters are to be defined in an LDA algorithm: number of topics, \textit{alpha}, and \textit{beta}.

The number of topics defines the latent topics to be extracted from the corpus. The parameter \textit{alpha} is \textit{a priori} belief in document-topic distribution, while \textit{beta} is \textit{a priori} belief in topic-word distribution.

To define the number of topics for each micro category corpus, we tried numbers from 1 to 5, constantly checking which configuration would result in the best average topic coherence for that corpus. Small corpora would have 1 or 2 topics, while bigger ones would have 5. To correctly define the \textit{alpha} and \textit{beta} priors, we would have to analyze the distribution for each category corpus \cite{wallach2009rethinking}. Since this would be rather difficult, we set those priors to be auto-defined by the LDA algorithm, which learns these parameters based on the corpus. We select the terms with the highest coherence with the resulting topics. Each topic returns 20 words with their coherence scores, but we do not use all of them as some have very low coherence. After testing a few configurations, for each topic, we select 60\% of the terms with the highest coherence within that topic. 

With the terms for each topic taxonomy, we ask an LLM to separate the ones closely related to the main topic from those unrelated, as mentioned earlier.

\subsection{Hierarchy Construction}

Once we have the post-processed lists of candidate terms obtained by each technique mentioned in subsection \ref{taxonomyC}, we merge them and remove repetitions. After the merge, for each macro category, we have lists of terms for each micro category, representing each topic taxonomy. However, they do not have any hierarchy level between the terms configuring the taxonomy.

To tackle this problem, we use an LLM again, this time with a prompt that searches for sub-categories within the terms of a topic to create these hierarchies. The prompt is illustrated below:

\begin{lstlisting}[caption=Prompt for creating a hierarchy for each list of tags. ,captionpos=b]
prompt="Create a dictionary by hierarchically arranging the following words:" + <words_list> +." Use JSON format as the output such as the following: {\"key\": [\" list of words\"]}"
\end{lstlisting}

With this prompt, we seek to ensure that the LLM response has a consistent pattern and facilitates handling the returned string. After this step, we have a hierarchy of terms in each topic taxonomy in the \emph{Food} and \emph{Shopping} macro categories.

\subsection{Merchant Tagging}
\begin{figure}[!ht]
\centering
\includegraphics[width=.99\columnwidth]{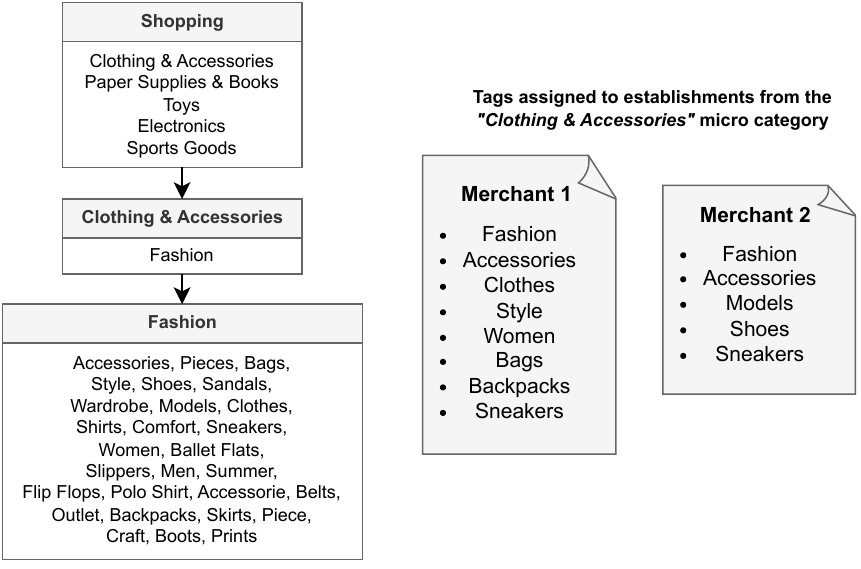}
\caption{Assigning tags to establishments based on a topic taxonomy.}
\label{fig:pipeline}
\vspace{-0.5em}
\end{figure}

With the topic taxonomies for both \textit{Food} and \textit{Shopping} macro-categories,  we can now assign tags to merchants/establishments. To do so, we use the descriptions attached to these establishments, and we see which terms from a taxonomy are mentioned in their descriptions with a reverse index algorithm. We employ the taxonomy whose topic is the same as the establishment's micro category, as shown in Figure \ref{fig:pipeline}.

\section{Taxonomy Evaluation}

To properly evaluate the topic taxonomies that we created in this work. We developed a two-step qualitative evaluation of a limited part of the results.

In total, 58 topic taxonomies were created for the \emph{Food} set and 6 for the \emph{Shopping} set. For our evaluation, we selected the topic taxonomies with the highest number of terms in each part (the "Brazilian Cuisine" taxonomy for the \emph{Food} part and the "Clothing and Accessories" taxonomy for the \emph{Shopping} one). First, we assess the quality of the taxonomy created, and then, we evaluate the tags assigned to establishments based on that taxonomy. We asked 12 volunteers to answer a two-part form.

\textit{Part 1 - Human Evaluation of the coherence of the topic taxonomy} - In this part, the volunteers were shown a taxonomy and asked to select which terms (if any) were irrelevant or weakly related to the main topic. This way, we can calculate the coherence of the topic by dividing the number of relevant terms per the total number of terms in that topic, similar to how \citealp{taxocom} evaluated TaxoCom. Both taxonomies are primarily coherent, according to our evaluators, with average coherence scores above 90\%.



\textit{Part 2 - Human Evaluation of the Quality of the Tagging Process} - 
In this part, the volunteers were asked to examine if the tags assigned to an establishment were appropriate and coherent to that establishment's description. We selected the top 5 establishments with the highest transactions for each micro category. We asked our evaluators to analyze the tags assigned to describe that establishment and choose the ones that were not appropriate. This way, we have a coherence ratio for each establishment based on the number of proper tags divided by the total number of tags. We average the results of our 12 evaluators and present them in Table \ref{table:eval2}. Figure \ref{fig:pipeline} shows the "Clothing \& Accessories" taxonomy that was evaluated and 2 of the merchants and the tags assigned to them that were included in the evaluation.

\begin{table*}[!htb]
\centering
{\footnotesize
\begin{tabular}{c|c|c|c|c|}
\cline{2-5}
                                         & \multicolumn{2}{c|}{\textbf{Brazilian Cuisine Taxonomy}} & \multicolumn{2}{c|}{\textbf{Clothing \& Accessories Taxonomy}} \\
                                         & \textbf{Average Coherence}   & \textbf{Number of Tags}   & \textbf{Average Coherence}      & \textbf{Number of Tags}     \\ \hline
\multicolumn{1}{|c|}{\textbf{Merchant 1}} & 92.30\%                      & 10                        & 97.11\%                         & 8                           \\ \hline
\multicolumn{1}{|c|}{\textbf{Merchant 2}} & 94.23\%                      & 8                         & 83.07\%                         & 5                           \\ \hline
\multicolumn{1}{|c|}{\textbf{Merchant 3}} & 89.23\%                      & 5                         & 94.38\%                         & 5                           \\ \hline
\multicolumn{1}{|c|}{\textbf{Merchant 4}} & 87.17\%                      & 6                         & 93.84\%                         & 5                           \\ \hline
\multicolumn{1}{|c|}{\textbf{Merchant 5}} & 93.40\%                      & 7                         & 97.43\%                         & 6                           \\ \hline
\end{tabular}}
\caption{Results of evaluating the tags assigned to each merchant/establishment.}
\label{table:eval2}
\end{table*}


\section{Taxonomy Expansion}
Another essential part of our method is the automatic expansion of existing taxonomies as new terms arrive. In this section, we present our approach to taxonomy expansion by using instruction-based LLMs. To our knowledge, this is the first work to use instructions in LLM prompting for taxonomy expansion.

\subsection{Prompt engineering instruction for taxonomy representation}

\begin{lstlisting}[caption=Prompt for representation of taxonomy,captionpos=b]
Childs of [ROOT]: [CHILD1,CHILD2,CHILD3] 
Childs of [CHILD1]: [CHILD4,CHILD5] 
Childs of [CHILD2]: [CHILD6] 
...
\end{lstlisting}

First, we must represent our topic taxonomies so that an LLM can understand their structure. We used the above generic prompt in all tested methods to convert topics to roots and their terms to child nodes.

\subsection{Predicting the parent of a node}

In our experiments, we used two datasets: our \textit{Food} and \textit{Shopping} topic taxonomies and the taxonomies from SemEval-2015 Task 17 \cite{popescu2015semeval}. Those are low-resource taxonomies, with thousands of nodes or less, which are appropriate for the current prompt size of LLMs. We used the SemEval dataset to compare the results with well-established methods for taxonomy expansion, such as Musubu \cite{musubu}. Similar to their experiments, we hid 20\% of the terms in the taxonomies to predict their respective parent nodes. To verify the parent/root of a new term, we used the following prompt:

\begin{lstlisting}[caption=Prompt for searching for a node's parent=b]
prompt="Who is the father of "+<new_term>+"?"
\end{lstlisting}

In table \ref{tab:avaliacao3}, we see the f1-Score for parent node prediction using baselines, such as Bert-Base and Musubu; Commercial LLMs, such as Gemini and GPT-4; And open-source LLMs, such as LLama-Alpaca(7B), Phi-2, and Mixtral 8x7B. We evaluate them in 4 taxonomies from the SemEval dataset and our taxonomies. For each taxonomy, the LLMs perform significantly better than Musubu, with commercial LLMs Gemini and GPT-4 having the highest f1-Scores, with the latter beating the former by a few points. However, the most recent open-source options(Phi-2 and Mixtral 8x7B) are closing in performance.  

\begin{table*}[!htb]
    \centering
    {\footnotesize
    \begin{tabular}{|c|cccc|cc|} 
    \hline
        { \textbf{Method}}        &  &  \textbf{SemEval-2015 Task 17}&  &  &   \textbf{ Our taxonomies} & \\ 
                &  \textbf{Chemical} &  \textbf{Equipment}   &  \textbf{Food} &  \textbf{Science}  &\textbf{Food}&\textbf{Shopping}
                
                \\ 

        \hline
        \textbf{Gemini}  &  \textbf{0.68}       &  \textbf{0.80}    &  \textbf{0.91} & \textbf{0.72} & \textbf{0.89} & \textbf{0.73}\\
        \hline
        GPT-4  &  0.65       &  0.78    &  0.89 & 0.70 & 0.87 & 0.71\\
        \hline

        Mixtral-8x7B  &  0.59       &  0.63    &  0.80 & 0.57 & 0.74 & 0.60\\
        \hline
        
        Phi-2    &  0.56       &  0.52    &  0.68 & 0.56 & 0.64 & 0.54\\
        \hline
        
        LLama 7B         & 0.51       & 0.42    & 0.58 &0.46 &0.60 &0.49 \\
        \hline

        Musubu & 0.35       & 0.46    & 0.37 &0.42 &0.21 &0.13 \\
        \hline
        
        Bert-Base  & 0.13       & 0.14    & 0.12 &0.16 &0.11 &0.06 \\
        \hline

    \end{tabular}}
\caption{F1-score for parent node prediction.}
    \label{tab:avaliacao3}
\end{table*}

It is important to note that while SemEval taxonomies have thousands of nodes, ours have only a few hundred, which we can assume is a significant reason for the degrading performance of Musubu and Bert (LMs or LM-based methods). In contrast, the LLMs have a robust performance in such low-resource settings. This also shows that LLMs have a remarkable understanding of questions and zero-shot performance, generalizing well even for datasets in different languages.

\section{Conclusion}
In this work, we presented an unsupervised method for automatically creating and expanding topic taxonomies using LLMs. We evaluated some of the generated taxonomies and applied them in transaction tagging in a retailer's bank dataset. The evaluation showed promising results, with average coherence scores above 90\% for the two selected taxonomies. The taxonomies' expansion with Gemini also showed exciting results for parent node prediction, with F1-scores of 89\% and 70\% for \emph{Food} and \emph{Shopping} taxonomies, respectively. 


For future works regarding the taxonomy construction part, we could test more robust methods for term selection, such as embedding-based ones. For the taxonomy expansion, we see many possible tasks that can be explored, such as node-level operations to sub-trees level, where we generate entire sub-trees and search for similar ones. We also intend to train our instruction-tuned LLM for taxonomy tasks by finetuning or using more efficient methods such as Lora \cite{hu2021lora}.

\section*{Limitations}
To discuss the limitations of our work, let us start with the Taxonomy Construction part. For this part of our method, we relied on Topic Modeling to select candidate terms for our taxonomies. The LDA algorithm for Topic Modeling performs poorly when the base corpus is small. For some of our topics, the corpora are pretty small, with less than 100 words in the vocabulary. This can lead to topics with irrelevant, not coherent terms. Also, we could have experimented more with the LDA hyperparameters for each micro-category corpora. 

For the evaluation part of the generated taxonomies, we did not include an evaluation for topic completeness. Since we have no ground truth, we cannot quantify how completely the terms in a taxonomy cover the main topic. Moreover, we only evaluated 2 of the 64 taxonomies we generated with our method, leaving much to analyze.

We only evaluated in a low-resource taxonomy setting in the expansion experiment, with less than a thousand nodes. Most studies are focused on taxonomies with hundreds of thousands or even more nodes. This is a challenging setting for LLMs since they have a limited context; other works that deal with this contextual limitation can be helpful to mitigate this problem \cite{liang2023unleashing}.

\section*{Ethics Statement}
In this work, we ensure the utmost protection of customers and store sensitive data by exclusively using non-sensitive information in our dataset. Our prompts solely rely on selected words from store descriptions, thus avoiding any direct usage of personal or sensitive information. No customer-specific data or store-sensitive details are integrated into the system, upholding privacy and security as top priorities.

Moreover, we strictly adhere to ethical guidelines during our experiments involving volunteers, and no personal data is collected from them. Our focus lies solely on analyzing the results of our proposed approach. Participants' anonymity and confidentiality are maintained throughout the research process, ensuring a responsible and trustworthy approach to data handling.

\bibliography{anthology}
\bibliographystyle{acl_natbib}

\end{document}